\documentclass[10pt,journal,compsoc]{IEEEtran}
\usepackage{amsmath,amsfonts}
\usepackage{algorithmic}
\usepackage{algorithm}
\usepackage{array}
\usepackage[caption=false,font=normalsize,labelfont=sf,textfont=sf]{subfig}
\usepackage{textcomp}
\usepackage{stfloats}
\usepackage{url}
\usepackage{multirow}
\usepackage{verbatim}
\usepackage{amssymb}
\usepackage{graphicx}
\usepackage{cite}
\usepackage{xcolor}
\usepackage{threeparttable}
\usepackage{mathtools}
\usepackage{booktabs}

\MHInternalSyntaxOn\makeatletter
\renewcommand{\xleftrightarrow}[2][]{\ext@arrow 5599\MT_leftrightarrow_fill: {#1}{#2}}
\makeatother\MHInternalSyntaxOff

\begin{document}
\title{Hierarchy Flow For High-Fidelity \\Image-to-Image Translation}

\author{Weichen Fan$^\dag$,
        Jinghuan Chen$^\dag$,
        Ziwei Liu,~\IEEEmembership{Member,~IEEE,}

\IEEEcompsocitemizethanks{\IEEEcompsocthanksitem Weichen Fan, Jinghuan Chen and Ziwei Liu are with S-Lab, Nanyang Technological University.\protect\\
E-mail: fanweichen2383@gmail.com, chenjh1997@yahoo.com, ziwei.liu@ntu.edu.sg
\IEEEcompsocthanksitem $\dag$ Equal contribution.}
}

\markboth{Journal of \LaTeX\ Class Files,~Vol.~14, No.~8, August~2015}%
{Shell \MakeLowercase{\textit{et al.}}: Bare Demo of IEEEtran.cls for Computer Society Journals}

\IEEEtitleabstractindextext{%
\begin{abstract}
Image-to-image (I2I) translation comprises a wide spectrum of tasks. Here we divide this problem into three levels: strong-fidelity translation, normal-fidelity translation, and weak-fidelity translation, indicating the extent to which the content of the original image is preserved. Although existing methods achieve good performance in weak-fidelity translation, they fail to fully preserve the content in both strong- and normal-fidelity tasks, e.g. sim2real, style transfer and low-level vision. 
In this work, we propose \textit{Hierarchy Flow}, a novel flow-based model to achieve better content preservation during translation.
Specifically, \textit{1)} we first unveil the drawbacks of standard flow-based models when applied to I2I translation.
\textit{2)} Next, we propose a new design, namely hierarchical coupling for reversible feature transformation and multi-scale modeling, to constitute Hierarchy Flow.
\textit{3)} Finally, we present a dedicated aligned-style loss for a better trade-off between content preservation and stylization during translation.
Extensive experiments on a wide range of I2I translation benchmarks demonstrate that our approach achieves state-of-the-art performance, with convincing advantages in both strong- and normal-fidelity tasks.
Code and models will be at \url{https://github.com/WeichenFan/HierarchyFlow}.
\end{abstract}

}

\maketitle

\IEEEdisplaynontitleabstractindextext

%
\IEEEpeerreviewmaketitle

\begin{figure*}[t]
  \centering
  \includegraphics[width=1.0\linewidth]{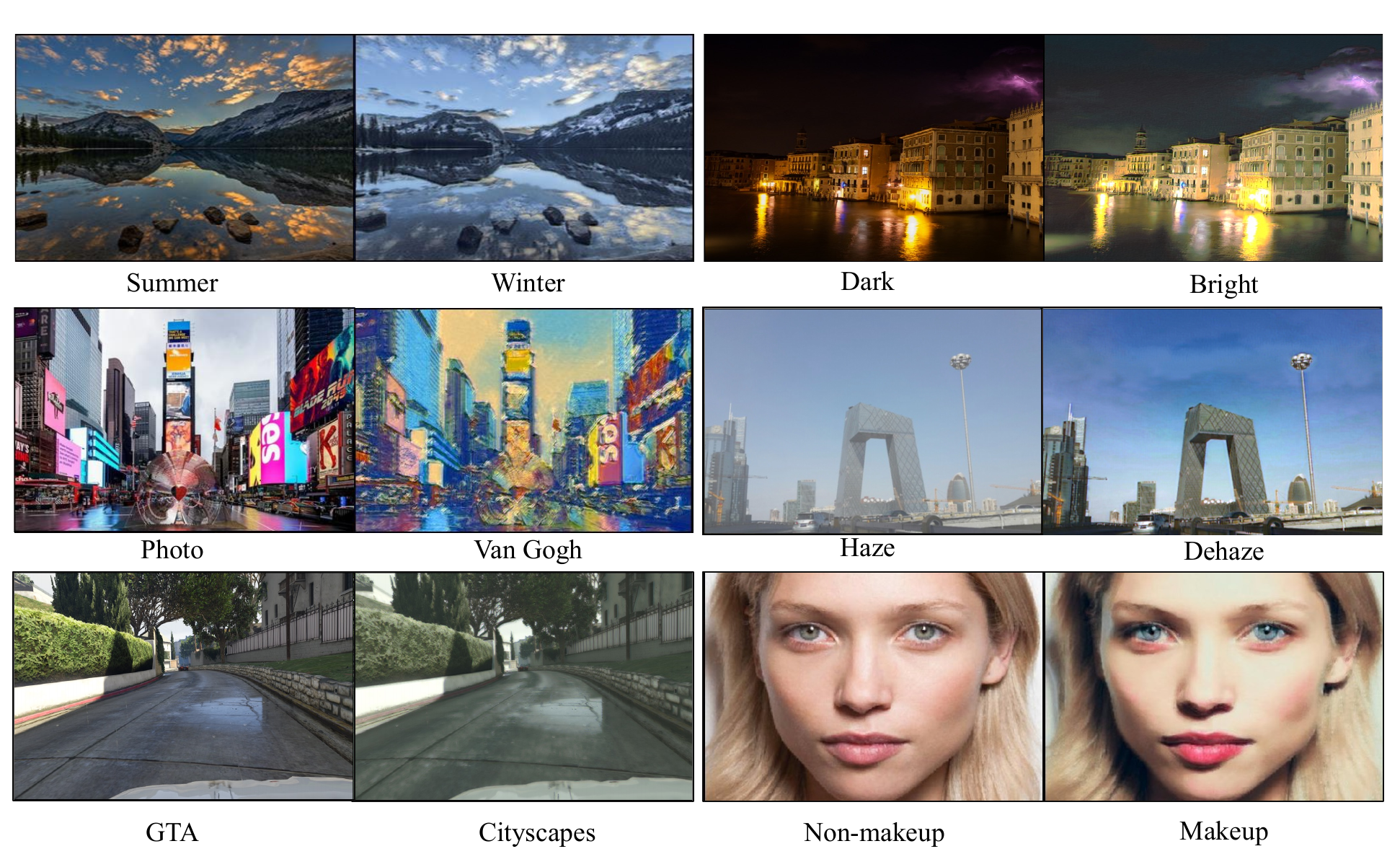}
  \caption{Given two images in different visual domains, our model learns to translate from one to the other with high fidelity in various tasks. (From top-left to bottom-right: 1. Summer to Winter \cite{zhu2017unpaired}; 2. Dark to Bright \cite{wei2018deep}; 3. Photo to Van Gogh \cite{wikiart}; 4. Hazy to Clear \cite{Dense-Haze_2019,NTIRE_Dehazing_2019}; 5. GTA \cite{richter2016playing} to Cityscapes \cite{cordts2016cityscapes}; 6. Non-makeup to makeup \cite{li2018beautygan}.}
  \label{fig:various_results}
\end{figure*}

\begin{figure}[t]
  \centering
  \includegraphics[width=1.0\linewidth]{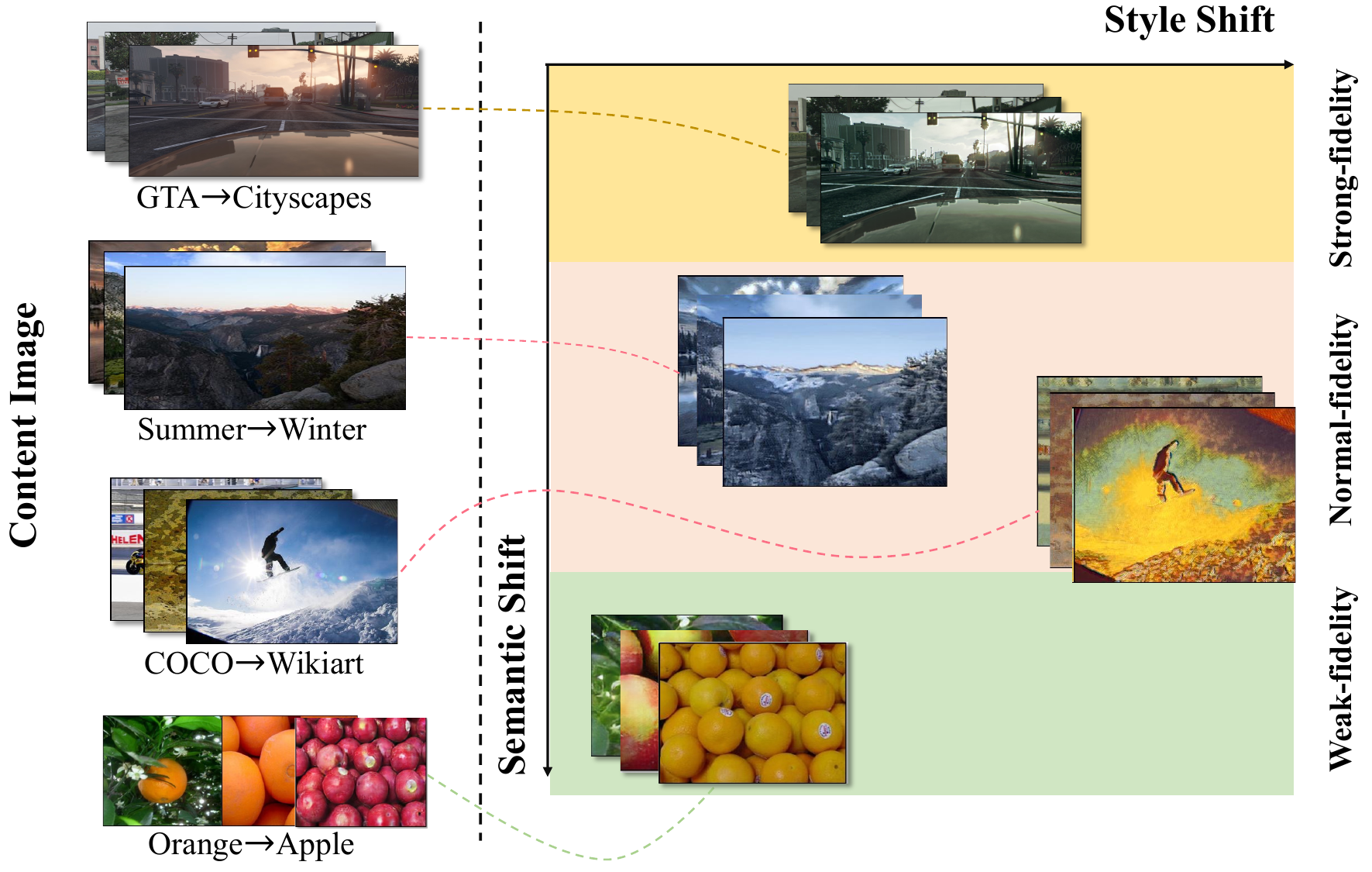}
  \caption{Illustration of three levels of image-to-image translation tasks: strong-, normal- and weak-fidelity translation, where the requirement of content preservation decreases gradually.
    }
  \label{fig:translation_subsets}
\end{figure}

\IEEEraisesectionheading{\section{Introduction}\label{sec:introduction}}
\IEEEPARstart{I}{mage-to-image} translation \cite{isola2017image} is a long-standing topic in computer vision, which is required to learn a mapping between two different visual domains while preserving the semantic information (content) of the source domain and obtaining the domain properties (style) of the target domain. Many applications, such as neural style transfer \cite{gatys2015texture,Gatys_2016_CVPR,huang2017arbitrary,li2017universal}, super-resolution \cite{Ledig_2017_CVPR,Lai_2017_CVPR}, image enhancement \cite{lore2017llnet} and photo-realistic synthesis \cite{Chen_2017_ICCV,Wang_2018_CVPR,richter2021enhancing}, can be formulated as I2I translation problems. Among most tasks, fully preserving semantic information during translation is important yet challenging, especially in scenarios where the content gap between source and target domains is large. According to the requirement of content preservation during translation, we further divide these tasks into three levels: strong-fidelity translation, normal-fidelity translation, and weak-fidelity translation (see Figure \ref{fig:translation_subsets}). In this work, we are interested in strong- and normal-fidelity settings, where content preservation plays a crucial role during translation.

Existing I2I translation methods can be broadly categorized into two approaches. Some methods learn a bijective mapping between source and target images, by forcing the translated images to be reconstructed back to the source images during training using a cyclic loss (e.g.  \cite{zhu2017unpaired}). Others try to fully disentangle content and style information from an image and achieve image translation by switching style information between source and target images (e.g.  \cite{zhu2017multimodal}). However, both approaches suffer from different levels of content distortion in translated images, since cyclic loss and feature disentanglement usually failed when rich and complex semantic information is required to be preserved, especially in strong- and normal-fidelity translation tasks. Relatively few methods focus on addressing this problem directly, and most among them suffer from carefully-designed tricks and are unable to generalize well to a wide range of tasks, since they either require auxiliary inputs such as paired images or additional information guidance, etc. (e.g.  \cite{richter2021enhancing}), or utilize complex contrastive training or pre-trained tasks aiming for better feature disentanglement learning (e.g. \cite{jia2021semantically,hoffman2018cycada}).

Content preservation remains a challenging problem in I2I translation. We consider flow-based models, also called normalizing flow, a subclass of deep generative networks that learns the exact likelihood of data distribution through a chain of basic blocks with fully-reversible transformations, which can be a perfect fit in the requirement of content preservation in image generation. ArtFlow \cite{an2021artflow} is the first work to use the flow-based model in I2I translation, specifically in style transfer task only. It proves the superiority of flow-based models in addressing the ``content leakage'' problem through lossless and unbiased feature extraction and image reconstruction. However, although ArtFlow achieves better content preservation compared to other methods, it suffers from severe checkerboard artifacts problem in the translated images (see Figure \ref{fig:check}). We further investigate the checkerboard issue and finally identify its root cause as the squeeze operation that is widely-used in flow-based models for multi-scale architecture  \cite{dinh2016density,kingma2018glow}. Sec.\ref{sec:preliminary} shows more analyses in detail. Therefore, we focus on designing a new framework that can utilize the superiority of flow-based models in content preservation for I2I translation, and also avoid the checkerboard artifacts problem as in ArtFlow.

In this work, we propose \textit{Hierarchy Flow}, which is a new flow-based model dedicated to unpaired I2I translation with good content-preserving ability. To avoid the problematic squeeze operation in multi-scale architecture for flow-based models, we present a novel basic block design, named \textit{Hierarchical Coupling Layer}, for efficient feature transformation and multi-scale modelling. In our model, feature extraction is done in a hierarchical way which can gradually remove style-specific features by a series of subtractive coupling operations in the forward pass. After feature extraction, we use Adaptive Instance Normalization \cite{huang2017arbitrary} to perform transformation upon deep features by replacing the statistical information (mean/std vector of features) of source features with those of target features. Finally, translated images are generated through the reversed pass of the network. Following most I2I translation tasks especially in style transfer \cite{huang2017arbitrary}, content loss and style loss calculated based on a pre-trained VGG encoder \cite{Simonyan2015very} are adopted. We further extend the idea of style loss and introduce its simple extension named \textit{aligned-style loss}, which takes the trade-off between content preservation and stylization into consideration, to further improve translation results, especially in scenarios where content gap is large between source and target domains, e.g. GTA \cite{richter2016playing} to Cityscapes \cite{cordts2016cityscapes} translation.

We apply the proposed framework to a wide range of applications, and plausible results (see Figure \ref{fig:various_results}) indicate the significance and effectiveness of the design in our method. To the best of our knowledge, we are the first I2I translation work that evaluates on both high-level (e.g. GTA to Cityscapes) and low-level (e.g. Low-light enhancement) vision tasks and achieves superior results in both areas. We summarize the contributions of this work as below: \textbf{1)} We divide image-to-image translation tasks into three subsets: strong-, normal- and weak-fidelity translation, according to the requirement of content preservation. \textbf{2)} We unveil the main drawback of flow-based models in I2I translation tasks, and propose Hierarchy Flow, a novel design for unpaired high-fidelity image-to-image translation. \textbf{3)} We design a novel aligned-style loss for efficient content-preserving feature transformation. \textbf{4)} We demonstrate that Hierarchy Flow outperforms previous methods with high fidelity and vivid stylization in extensive experiments.
\begin{figure}[t]
  \centering
  \includegraphics[width=0.6\linewidth]{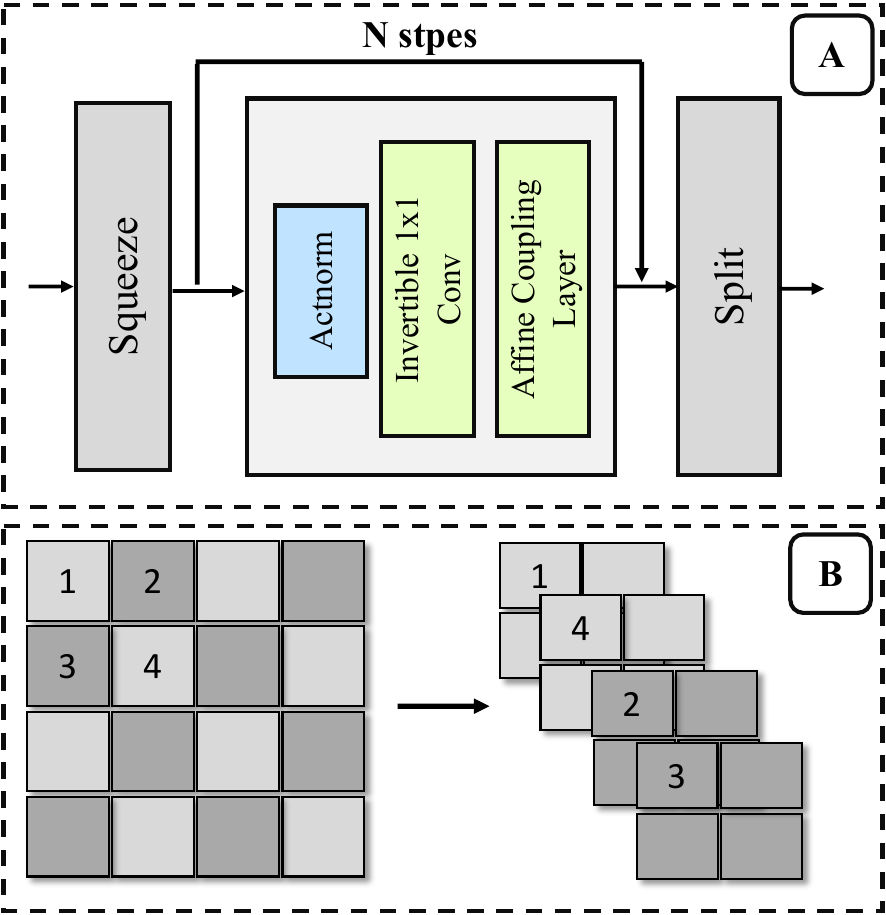}
  \caption{(A). The basic components of GLOW \cite{kingma2018glow} consist of a squeeze operation followed by a series of invertible layers for non-linear transformation. (B). The squeeze operation \cite{dinh2016density} reorganizes the features map following a spatial checkerboard pattern.}
  \label{fig:squeeze}
\end{figure}
\section{Related Work}
\label{sec:related}
\subsection{Image-to-Image Translation}
\noindent\textbf{Generic Image-to-Image Translation}: The previous generic I2I models \cite{ma2018exemplar,zhu2017multimodal,lee2018diverse} suffer from the problem of content distortion, even though a lot of regularization methods have been proposed to reduce the impact, including cyclic consistency, contrastive learning, etc. For weak-fidelity translations, where the content can be heavily modified, these methods are appropriate, while for strong- and normal-fidelity translations, they do not perform well. Our proposed model can be applied to both normal- and strong-fidelity translations without the problem of content distortion.\\[1ex]
\noindent\textbf{Strong-fidelity Translation}: Strong-fidelity I2I translation means the content of the source image should be preserved to a great extent. Sim2real translation \cite{richter2021enhancing,hoffman2018cycada}, colorization \cite{kumar2021colorization}, and low-level visions such as low light enhancement \cite{lore2017llnet,wei2018deep}, raindrop removal \cite{qian2018attentive,tu2022maxim} and dehazing \cite{engin2018cycle,anvari2020dehaze,tu2022maxim} belong to the strong-fidelity setting. These problems require the translated images to retain the exact rich and complex semantic information in source images. To achieve high content preservation, previous methods require paired training images or auxiliary inputs such as semantic segmentation masks. Most recently, VSAIT \cite{theiss2022unpaired} proposes a new framework based on vector symbolic architectures to directly solve ``semantic flipping'' problem and achieve current SOTA results in unpaired I2I translation. \\[1ex]
\noindent\textbf{Normal-fidelity Translation}: Normal-fidelity I2I translation includes style transfer \cite{gatys2015texture,huang2017arbitrary,chen2016fast,li2017universal,wu2022ccpl,an2021artflow}, season and weather transfer \cite{li2021weather}, etc. In this setting, the source and target domains usually show different visual effects, such as weather conditions and artistic styles, but share similar structural information, the primary objective is to transfer the overall visual effects of source domains to match those in target domains. Previous work have shown plausible overall visual results in these tasks, while a certain level of content distortion can be found when we zoom in to the details of translated images.\\[1ex]
\noindent\textbf{Weak-fidelity Translation}: Weak-fidelity I2I translation refers to problems where the source and target images may lie in completely different domains or modals, the translation is to be performed on a high semantic level, which means the content information of source images can be modified a lot. Label to image \cite{lin2018conditional} and object to object translation \cite{huang2018multimodal,lee2018diverse,zhu2017unpaired,park2020contrastive} belong to this type of problem.\\
\begin{figure}[t]
  \centering
  \includegraphics[width=1.0\linewidth]{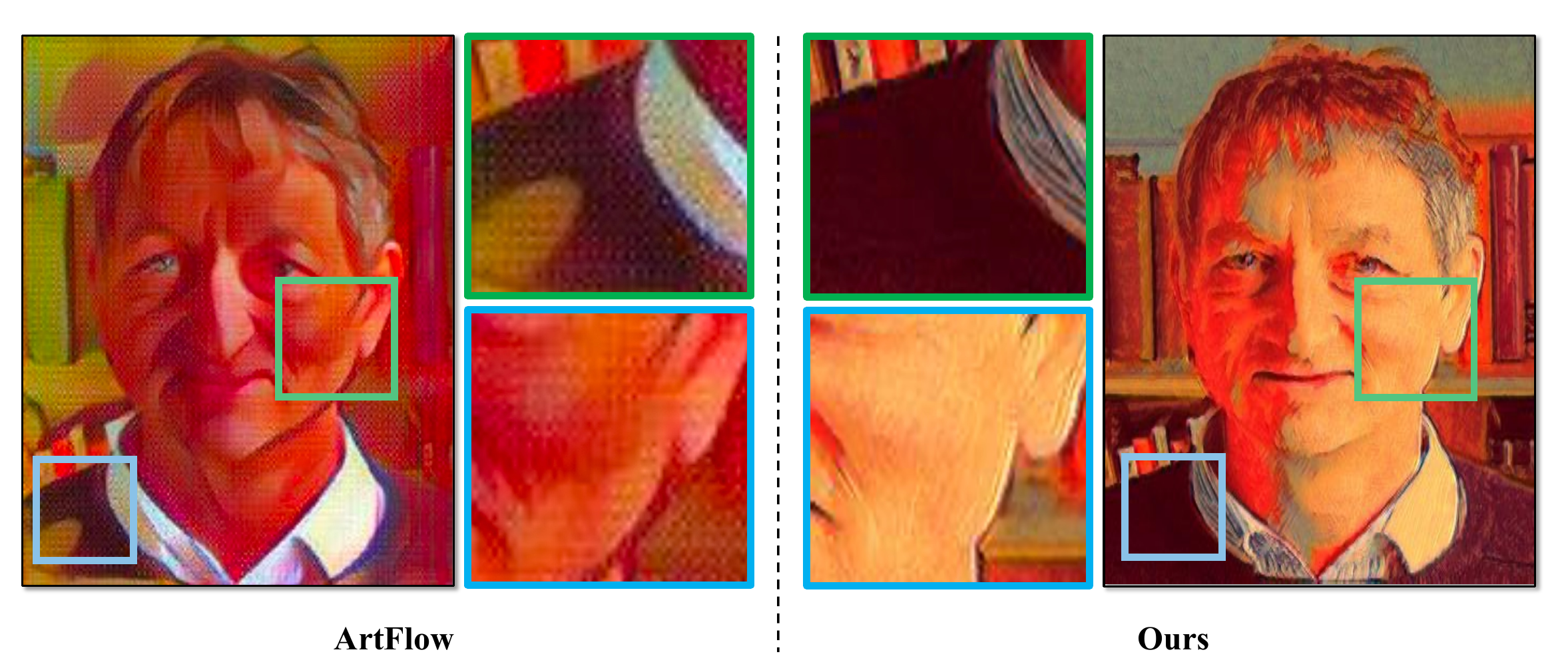}
  \caption{The checkerboard artifacts. Left: output sample from ArtFlow \cite{an2021artflow}; Right: our result. Compared to the output of ArtFlow which suffers from checkerboard artifacts (shown in zoomed-in boxes), we generate smooth images with high fidelity.}
  \label{fig:check}
\end{figure}
\subsection{Normalizing Flow}
Normalizing flow is a type of  generative model that uses a sequence of invertible mappings to transform from distribution to distribution, and it is accurate and efficient in both density estimation and sampling \cite{kobyzev2020normalizing}. Dinh et al. \cite{dinh2014nice} first propose a flow-based generative model, NICE. After that, GLOW \cite{kingma2018glow}, RealNVP \cite{dinh2016density}, and FLOW++ \cite{ho2019flow++} are proposed to improve the sample efficiency and density estimation performance. More recently, BeautyGLOW \cite{chen2019beautyglow} is proposed for makeup transfer. Besides, ArtFlow \cite{an2021artflow} proves that the normalizing flow is unbiased in neural style transfer compared with the previous work.

\section{Our Approach}
\label{sec:methodology}
In this section, we first give a brief introduction of flow-based generative models and unveil its main drawback in I2I translation, which is the checkerboard artifacts, in Sec.\ref{sec:preliminary}; next, we introduce the design of our proposed Hierarchy Flow in details in Sec.\ref{sec:approach}, which solves the checkerboard issues of previous methods and achieve better content preservation in high-fidelity image translation.
\subsection{Preliminary}
\label{sec:preliminary}
\subsubsection{Flow-based Generative Model}
Flow-based model is a subset of generative models that learns the exact log-likelihood of a high dimensional data distribution through a sequence of fully reversible transformations. Let $x$ be a high-dimensional variable with unknown distribution $x \sim p(x)$, a generative model $p_\theta(x)$ with parameters $\theta$ is designed to estimate distribution $p(x)$ given dataset $\mathcal{D}$, with training objective $min$ $\mathcal{L(D)}$, where 
\begin{align}
    \mathcal{L(D)} = \frac{1}{N}\sum_{i=1}^N -log(p_\theta(x^{(i)}))
\end{align}
Most flow-based generative models \cite{dinh2014nice,dinh2016density,kingma2018glow} formulate the generative process as
\begin{align}
    x \xleftrightarrow{f_1} h_1 \xleftrightarrow{f_2} h_2 \cdots \xleftrightarrow{f_k} z
\end{align}
where $z$ is latent variable, $f_\theta = f_1 \circ f_2 \circ \cdots \circ f_k$ is a sequence of invertible functions such that $z$ and $x$ satisfy the relationship $z=f_\theta(x)$ and $x=f_\theta^{-1}(z)$.
One of the most frequently-used flow-based networks for image synthesis is GLOW \cite{kingma2018glow}. As shown in Figure \ref{fig:squeeze}, it combines a series of steps of flow in a multi-scale architecture with squeeze operations \cite{dinh2016density} between each scale to effectively transform feature scale by trading spatial size for channel size. In each step of flow, features are transformed by actnorm activation followed by an invertible 1x1 convolution and lastly an affine coupling layer. Previous work ArtFlow \cite{an2021artflow} follows the network design of GLOW to achieve reversible transformations in universal style transfer, and solve the content leakage problem by lossless and unbiased image projection and reversion.

\begin{figure*}[t]
  \centering
  \includegraphics[width=1.0\linewidth]{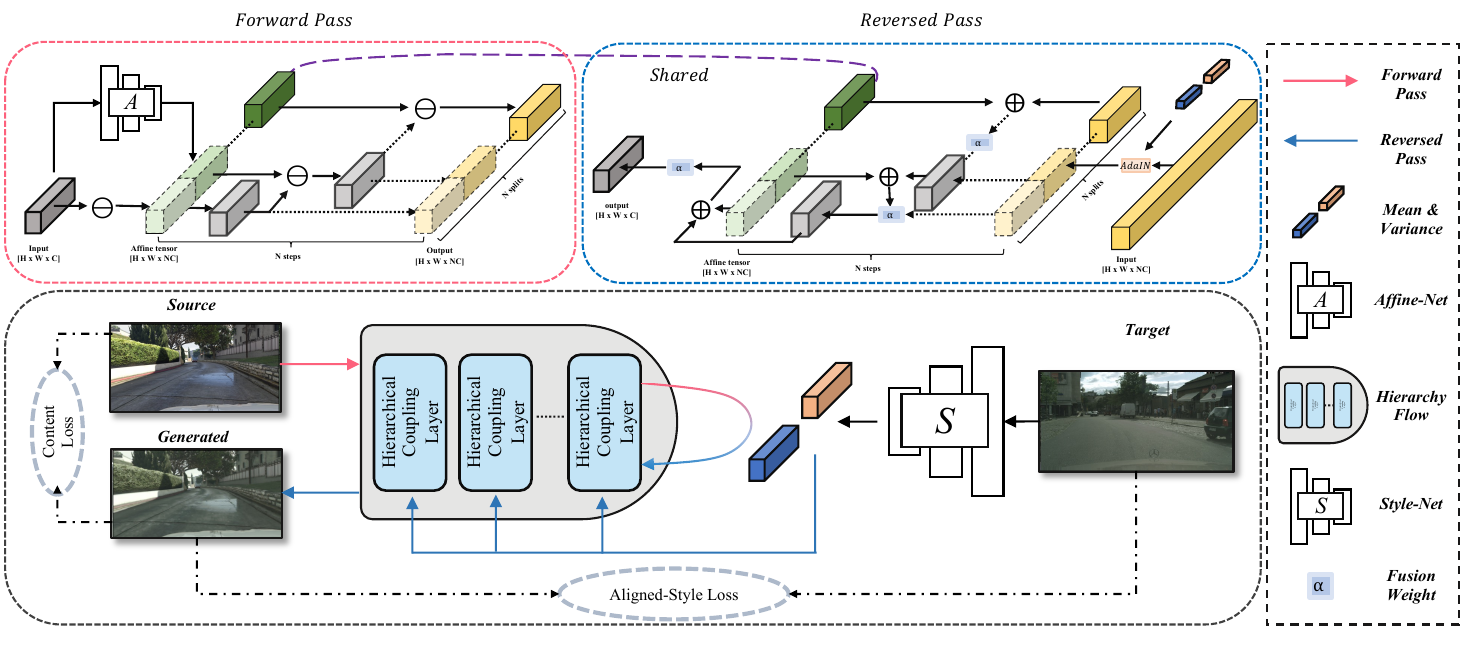}
  \caption{The framework of our proposed Hierarchy Flow. Given a source image and a target image, the source image is encoded by \textit{Hierarchy Flow} in the forward pass represented by the \textcolor{red}{red} arrows, then fused with style information extracted from the target image by \textit{Style-Net} in the form of AdaIN \cite{huang2017arbitrary}, and finally reverse-decoded by \textit{Hierarchy Flow} and denoted in \textcolor{blue}{blue} arrows to generate translated image.}
  \label{fig:overview}
\end{figure*}
\subsubsection{Checkerboard Artifacts Problem}
The squeeze operation plays a vital role in GLOW for multi-scale feature transformation. As shown in Figure \ref{fig:squeeze}(B), it follows a spatial checkerboard pattern to transform a $H\times W \times C$ tensor (left) into $\frac{H}{2}\times\frac{W}{2}\times4C$ tensor (right), to efficiently implement multi-scale architecture by trading spatial size for channel size. During reversed pass of the model, one can easily ``undo'' the squeeze operation by restoring the spatial dimension of the tensors. 

In ArtFlow, the style feature transfer module (AdaIN \cite{huang2017arbitrary} or WCT \cite{li2017universal}) is applied to the encoded features before the reversed pass to generate translated images. As a result, significant changes have been made to each individual channel of the features, and spatial misalignment will be produced when unsqueeze operations are performed on the transferred features during the reversed pass, leading to the obvious checkerboard artifacts in the output samples (see Figure \ref{fig:check}: Left). 

In order to utilize the usage of invertible network design of flow-based models and to solve checkerboard artifacts problem in ArtFlow, we aim to re-design the squeeze operation for multi-scale architecture to achieve content-fixed and artifacts-free image-to-image translation.

\subsection{Hierarchy Flow}
\label{sec:approach}
As shown in Figure \ref{fig:overview}, we propose Hierarchy Flow, which is a flow-based model with a novel design of basic block named \textit{Hierarchical Coupling Layer}. In general, given a set of images ($I_s$, $I_t$), a series of hierarchical coupling layers encode the source image $I_s$ to obtain the source features in the forward network inference. The target image $I_t$ is fed into a \textit{Style-Net} to obtain the style features. After that, we use AdaIN \cite{huang2017arbitrary} to perform style feature transfer to fuse the source features and style features, and finally perform image reconstruction through the reversed pass of the network to generate a translated image. In our model, the network architecture is carefully designed to be fully reversible. Therefore, combined with AdaIN, we can achieve I2I translation with desired content preservation.
\subsubsection{Hierarchical Coupling Layer}
By combining squeeze operation with affine coupling layer together, \textit{hierarchical coupling layer} enables complex feature transformation and multi-scale modeling inside one single block without spatial squeezing. Instead, we use hierarchical subtraction along the channel dimensions to implement spatial feature fusion and transformation in a learnable manner. Algorithms 1 and 2 show the details of the forward and reversed pass respectively.\\[1ex]
\textbf{A. Forward Pass.}
Given an input tensor $x$ with dimension $[H\times W\times C]$, we first apply an affine transformation with an \textit{Affine-Net} which can be any neural network, where the input tensor undergoes a channel-wise expansion and is mapped to an affined tensor with dimension $[H\times W\times nC]$ with expansion rate $n$. With the affine tensor separated into $n$ splits along channel dimension, we then apply a hierarchical subtractive coupling for $x$ in n-steps, and obtain output $y$ by concatenating the $n$ intermediate feature maps. \\[0.5ex]
\textbf{B. Reversed Pass.}
To compute the inverse of the above transformation, we can simply apply $n$ steps of  addictive coupling between input tensor $y$ and affine tensor $a$, and fuse the $n$ intermediate feature maps to obtain output tensor $x$. To better facilitate the fusion process in the training, we apply a learnable fusion weight $\alpha$ in each step of fusion, which measures the importance of each split of features during spatial fusion and transformation adaptively. \\[1ex]
\indent With the design of hierarchical coupling, we enable multi-scale feature transformation and fusion inside each basic block. Therefore, we can easily stack multiple blocks directly to implement more complex network modeling, without spatial squeezing as in previous flow-based methods, since adaptive spatial fusion has been applied inside each block. Despite of its simplicity in network design, Hierarchy Flow shows great improvement in high-fidelity translation tasks with better content preservation and artifacts-free image generation, which indicates the effectiveness and significance of our proposed method.

\begin{algorithm}[t]
\caption{Forward Pass.}\label{alg:alg1}
\begin{algorithmic}
\STATE 
\STATE {\textsc{FORWARD}}$(x)$
\STATE \hspace{0.5cm}$ a = \textbf{Affine-Net} (x)  $
\STATE \hspace{0.5cm}$ a_1,a_2,\cdots,a_n = \textbf{split}(a) $
\STATE \hspace{0.5cm}$ h_1 = x - a_1 $
\STATE \hspace{0.5cm}$ h_i = h_{\mathbf{i}-1} - a_\mathbf{i}  $ \textbf{ for } $ \mathbf{i}\gets 2,n $
\STATE \hspace{0.5cm}$ y = \textbf{concat}(h_1,h_2,\cdots,h_n) $
\STATE \hspace{0.5cm}\textbf{return}  $\textbf{y}$
\end{algorithmic}
\end{algorithm}

\begin{algorithm}[t]
\caption{Reversed Pass.}\label{alg:alg2}
\begin{algorithmic}
\STATE 
\STATE {\textsc{REVERSED}}($y$, $a_{1\cdots n}$, style feature $\mu$, $\sigma$)
\STATE \hspace{0.5cm}$ y = \textbf{AdaIN} (y,\mu,\sigma)  $
\STATE \hspace{0.5cm}$ y_1,y_2,\cdots,y_n = \textbf{split}(y) $
\STATE \hspace{0.5cm}$ h_n = y_n + a_n $
\STATE \hspace{0.5cm}$ h_{\mathbf{i}} = \alpha\cdot (y_\mathbf{i}+a_\mathbf{i}) + (1-\alpha)\cdot h_{\mathbf{i}+1} $ \textbf{ for } $ \mathbf{i}\gets n-1,1 $
\STATE \hspace{0.5cm}$ x = h_1 $
\STATE \hspace{0.5cm}\textbf{return}  $\textbf{x}$
\end{algorithmic}
\end{algorithm}

\subsubsection{Style-Net and AdaIN}
\label{sec:stylenet}
Our \textit{Style-Net} follows the design of Style-Encoder in MUNIT \cite{huang2018multimodal}, which consists of a series of convolutional layers with stride 2 followed by a global average pooling and two linear layers that output a mean vector $\mu$ and a variance vector $\sigma$. The purpose of the style network is to extract style information, thus we do not use any normalization layers in the network, which would modify the style information. 

AdaIN is first proposed in  \cite{huang2017arbitrary}, it separates deep features into normalized feature map and mean/std vectors, which can be
referred as content and style information respectively. To perform style feature transfer, it first scales normalized source feature $x$ by variance vector $\sigma$, then shift it with mean vector of $\mu$, where $\sigma$ and $\mu$ are the outputs of \textit{Style-Net}:
\begin{align}
    \text{AdaIN}(x, \mu, \sigma) = \sigma(\frac{x-\mu(x)}{\sigma(x)})+\mu
\end{align}
\noindent In our model, AdaIN is applied to every hierarchical coupling layer before the reversed pass.
\begin{figure*}[t]
    \centering
    \includegraphics[width=1.0\linewidth]{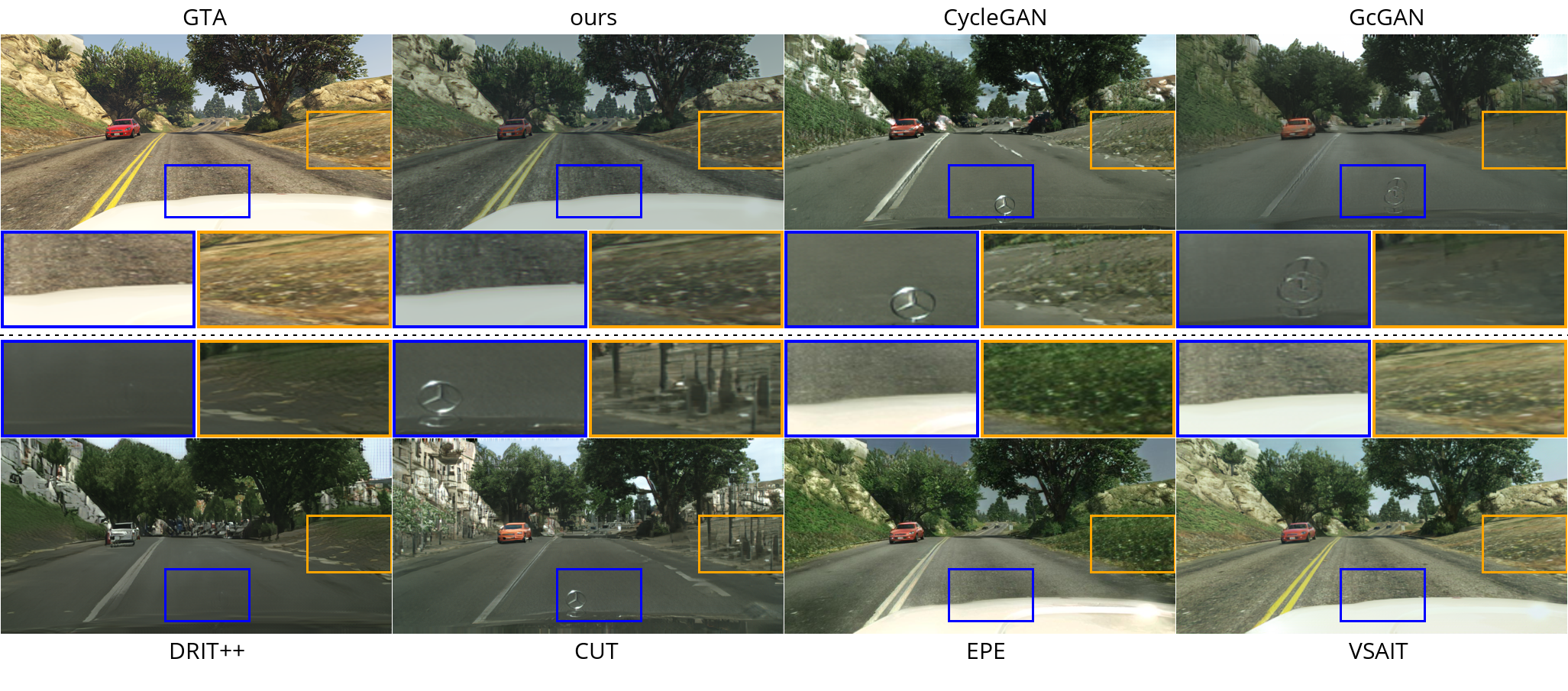}
    \caption{Visual results of GTA to Cityscapes compared with the state-of-the-art methods. CycleGAN \cite{zhu2017unpaired}, GcGAN \cite{fu2019geometry} and CUT \cite{park2020contrastive} hallucinate logos (\textcolor{blue}{blue} box), while DRIT++ \cite{lee2020drit++} and EPE \cite{richter2021enhancing} generate extra grass on the ground (\textcolor{orange}{orange} box). Compared to VSAIT \cite{theiss2022unpaired}, our model generates better Cityscapes-style images.}
    \label{fig:GTA_Figure}
\end{figure*}
\subsubsection{Loss Function}
\label{sec:Loss}
Our objective function can be expressed as:
\begin{align}
      \mathcal{L} = \mathcal{L}_c + \lambda \mathcal{L}_{as}
\end{align}
where $L_c$ is the content loss, $L_{as}$ is our proposed aligned-style loss, and $\lambda$ is a weighting factor used to trade-off between content and style. \\[0.5ex]
\noindent\textbf{Content Loss}:
Following  \cite{huang2017arbitrary}, the content loss is defined as the Euclidean distance between the channel-wise normalization of VGG features for the generated image $\hat{x}$ and the source image $x$.
\begin{align}
      \mathcal{L}_{c} = \left \|norm(\phi(\hat{x})) - norm(\phi(x))\right \|_2
\end{align}
where $\phi$ refers to the layer $relu\ 4\_1$ of a pre-trained VGG-19 encoder, $norm$ denotes the channel-wise normalization.\\[0.5ex]
\noindent\textbf{Aligned-Style Loss}:
Considering that the semantic information extracted from VGG-19 of the source image and the target image are not exactly matched in unpaired translation, we extend the style loss in  \cite{huang2017arbitrary} to be \textit{aligned-style loss} by setting a parameter $k$ to adjust the percentage of extracted tensors that are used for loss computation. We define $S$ as an ascending sort function. Given a source image $x$, a target image $y$, and the transferred image $\hat{x}$, with an energy function $E(\phi_i(\hat{x}),\phi_i(y))=\left \|\mu(\phi_i(\hat{x}))-\mu(\phi_i(y)) \right \|_2$, where $\phi_i$ ($i\in L=\{1,2,3\}$) represents a set of pre-trained VGG-19 layers $\{relu1\_1,relu2\_1,relu3\_1\}$, we could have the chosen index:
\begin{align}
      \mathcal{C}= \{c \in \mathbb{N}_{S}|c\leqslant kN,0<k\leqslant 1 \}
\end{align}
where $\mathbb{N}_{S}$ is the set of indexes of the sorted tensor $S(E(\phi_i(\hat{x}),\phi_i(y)))$, $N$ denotes its total length, and $k$ is the weighting parameter. Therefore,
\begin{align}
      \mathcal{L}_{as} = \sum_{i=1}^L\sum_{j\in C}\left \|\mu(\phi_i(\hat{x})_j) - \mu(\phi_i(y)_j)\right \|_2 + \\ \nonumber \sum_{i=1}^L\sum_{j\in C}\left \|\sigma(\phi_i(\hat{x})_j) - \sigma(\phi_i(y)_j)\right \|_2
\end{align}
where $\phi_i(x)_j$ denotes the $j^{th}$ channel of the output tensor of the $i^{th}$ layer from the set  $\{relu1\_1,relu2\_1,relu3\_1\}$ of a pre-trained VGG-19 encoder.

\begin{table*}[!t]
\caption{Quantitative evaluation on GTA to Cityscapes. Metrics include average pixel prediction accuracy (pxAcc), average class prediction accuracy (clsAcc), mean IoU (mIoU), and SSIM and FID. We achieve best results with smallest model size.\label{tab:gta2city}}
\centering
\resizebox{0.79\textwidth}{!}
{
\begin{tabular}{c|c|c|c|c|c|c|c}
\toprule[1.2pt]
 Method & pxAcc$\uparrow$ & clsAcc$\uparrow$ & mIoU$\uparrow$ & SSIM$\uparrow$ & FID$\downarrow$ &Params$\downarrow$ & FLOPs$\downarrow$ \\
\midrule
CycleGAN \cite{zhu2017unpaired} & 75.93 & 39.12 & 28.92 & 0.70 & 22.12 & 22.76M & 454.91G \\
GcGAN \cite{fu2019geometry} & 70.26 & 38.13 & 27.34 & 0.67 & 12.32 & 7.84M & \underline{84.71G} \\
DRIT++ \cite{lee2020drit++} & 75.89 & 35.44 & 27.31 & 0.55 & 14.69 & 17.08M & 170.18G \\
CUT \cite{park2020contrastive} & 70.81 & 37.12 & 26.43 & 0.61 & 21.18 & 11.38M & 128.26G \\
ArtFlow \cite{an2021artflow} & \underline{76.05} & \underline{45.37} & \underline{32.15} & 0.52 & 8.59 & \underline{6.42M} & 90.2G \\
VSAIT \cite{theiss2022unpaired} & 75.33 & 42.23 & 30.33 & \textbf{0.93} & 7.94 & 11.38M & 128.26G\\
\textbf{ours} & \textbf{79.63} & \textbf{45.67} & \textbf{33.76} & \underline{0.87} & \textbf{7.87} & \textbf{0.68M} & \textbf{10.13G} \\
\bottomrule[1.2pt]
\end{tabular}}
\end{table*}

\section{Experiments}
To demonstrate the effectiveness of our method in normal- and strong-fidelity translation tasks, we show comparisons between our proposed Hierarchy Flow and other state-of-the-art methods of respective fields in this section. More results can be found in supplementary materials. 
\subsection{Experimental Setup}
\noindent\textbf{Network Architecture.} As mentioned in Sec \ref{sec:approach}, a stacked sequence of basic blocks can be used for complex modeling in different tasks. In the following experiments, we introduce 4 variants of model size with different number of basic blocks for different tasks, which includes:\\[0.5ex]
\indent1) \textit{HF}: 2 blocks with expansion rates $[10,4]$ (Sec.~\ref{sec:GTA}); \\[0.5ex]
\indent2) \textit{HF+}: 3 blocks with expansion rates $[4,5,2]$ (Sec.~\ref{sec:style_trans}); \\[0.5ex]
\indent3) \textit{HF++}: 3 blocks with expansion rates $[10,4,4]$; \\[0.5ex]
\indent4) \textit{$\text{HF}^\dag$}: 4 blocks with expansion rates $[10,4,4,4]$ (Sec.~\ref{sec:llv}). \\
More studies of the network structure can be found in the ablation study. \\[1ex]
\noindent\textbf{Affine-Net Design.} Inside each basic block, \textit{Affine-Net} is defined as a 3-layer perceptron with \textit{Conv-IN-ReLU-Conv-IN-ReLU-Conv-ReLU} in specific, where the first two convolutional layers double the input channel dimension and the last one maps feature to output channel dimension respectively. All 3 conv layers are designed with 3x3 kernels and stride 1.

\noindent\textbf{Implementation Details.} We implement Hierarchy Flow in the Pytorch framework, and train for 300k iterations using an Adam optimizer with a batch size of 1, an initial learning rate of 1e-5, and a cosine annealing scheduler which continuously decreases the learning rate to 0. The loss weight $\lambda$ is set to 0.1, and $k$ in aligned-style loss is set to 0.8 unless specified. In each experiment, we train the model with 10 random seeds and report the average quantitative results among them.

\subsection{GTA to Cityscapes [Strong-Fidelity]}
\label{sec:GTA}
\noindent\textbf{Dataset}:
We use GTA dataset \cite{richter2016playing} as the source domain and Cityscapes \cite{cordts2016cityscapes} as the target domain. By default, all images are resized to $512\times256$ and randomly cropped to $256\times256$ during training. Evaluation is performed in $512\times256$.\\[1ex]
\noindent\textbf{Qualitative Evaluation}:
For the comparison to the previous work, we select several generic I2I models that focus on better semantic alignment and a specific photo-realism model EPE \cite{richter2021enhancing} trained with auxiliary inputs. As shown in Figure \ref{fig:GTA_Figure}, except VSAIT \cite{theiss2022unpaired}, all previous I2I models fail to retain full content information, and different levels of content distortion exist. More details are shown in the \textcolor{blue}{blue} box and \textcolor{orange}{orange} box in Figure \ref{fig:GTA_Figure}).\\[1ex]

\begin{table}[!t]
\caption{Quantitative results of art-style transfer. ArtFlow with $*$ and $\dag$ represents the combination with AdaIN and WCT respectively. SSIM and KID($\times {10}^3$) are used as metrics. Our model achieved competitive results on both stylization and content preservation with the lowest parameter number and FLOPs.\label{tab:style_res}}
\centering
{
\begin{tabular}{c|c|c|c|c}
\toprule[1.2pt]
Method & SSIM$\uparrow$ & KID$\downarrow$ & Params$\downarrow$ & FLOPs$\downarrow$ \\
\midrule
AdaIN \cite{huang2017arbitrary} & 0.28 & 41.1/5.1 & 7.01M & 117.5G\\
WCT \cite{li2017universal} & 0.24 & 51.2/6.2 & 34.24M & 272.3G \\
ArtFlow* \cite{an2021artflow} & 0.52 & \textbf{24.6/3.8} & \underline{6.42M} & 105.2G\\
ArtFlow$\dag$ \cite{an2021artflow} & \underline{0.53} & 33.3/5.3 & \underline{6.42M} & 105.2G\\
CCPL \cite{wu2022ccpl} & 0.43 & 39.1/6.8 & 8.67M & \underline{90.2G}\\
\textbf{ours} & \textbf{0.60} & \underline{28.2/4.7} & \textbf{1.01M} & \textbf{24.6G} \\
\bottomrule[1.2pt]
\end{tabular}}
\end{table}

\noindent\textbf{Quantitative Evaluation}:
To quantitatively evaluate the performance of GTA to Cityscapes, it is critical to choose suitable metrics that can measure the ability of content preservation during translation. As illustrated in  \cite{jia2021semantically}, popular metrics like FID and KID ignore semantic mismatch during evaluation and thus can be misleading. Instead, in GTA to Cityscapes translation, we can utilize the semantic correspondence between images and segmentation labels as a reference during evaluation. Specifically, for each method, we use a lightweight DeepLabV3 \cite{chen2017rethinking} model to train on translated GTA images and segmentation masks and report the semantic segmentation evaluation on the validation set of Cityscapes, which reflects the performance of both content preservation and stylization at the same time.
Additionally, we report the Structural Similarity Index Measure (SSIM) between translated images and source images, which also measures the performance of content preservation. 
Since EPE uses additional ``G-buffers'' information which is not publicly available to reproduce their method, the quantitative result of EPE is omitted.
As shown in Table \ref{tab:gta2city}, our model outperforms previous methods by a large margin including the current SOTA method VSAIT.
\begin{table}[!t]
\caption{Human preference score. ``Detail'' and ``Overall'' denote the evaluation criteria of content preservation and overall performance. Our model surpass the previous SOTA methods CCPL \cite{wu2022ccpl} by \textbf{68.1\%} in ``Detail'' and \textbf{30.3\%} in ``Overall''.\label{tab:user}}
\centering
\resizebox{0.49\textwidth}{!}
{
\begin{tabular}{c|c|c|c|c|c}
\toprule[1.2pt]
Method & \textbf{Ours} & CCPL \cite{wu2022ccpl} & ArtFlow \cite{an2021artflow} & AdaIN \cite{huang2017arbitrary} & WCT \cite{li2017universal}\\
\midrule
Detail$\uparrow$ & \textbf{76.5\%} & 8.40\% & \underline{15.1\%} & 0\% & 0\%\\
Overall$\uparrow$ & \textbf{47.9\%} & 17.6\% & \underline{25.2\%} & 5.9\% & 3.3\%\\
\bottomrule[1.2pt]
\end{tabular}}
\end{table}

\begin{figure*}[t]
  \centering
  \includegraphics[width=1.0\linewidth]{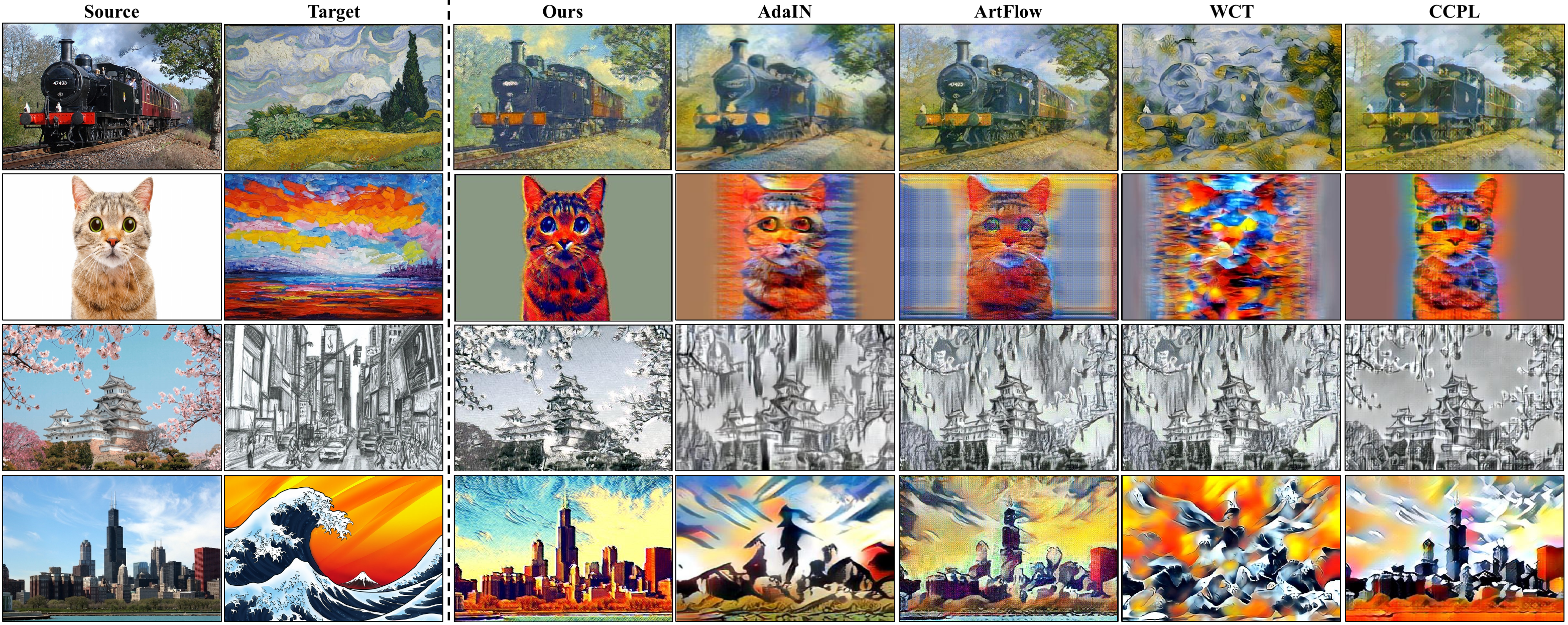}
  \caption{Style transfer results compared with the state-of-the-art style transfer methods. Compared to other methods, our model generates images with satisfying artistic style without losing content information.
   }
  \label{fig:art_pic}
\end{figure*}
\subsection{Artistic Style Transfer [Normal-Fidelity]}
\label{sec:style_trans}
\noindent\textbf{Dataset}: Following previous artistic style transfer work, we use MS-COCO \cite{lin2014microsoft} as source domain and Wiki-Art \cite{wikiart} as target domain in our experiments. By default, all images are resized to $300\times400$ for training and testing.\\[1ex]
\noindent\textbf{Qualitative Evaluation}:
We compare the visual performance with different methods. WCT \cite{li2017universal} generates stylized images with severe content distortion. AdaIN \cite{huang2017arbitrary} preserves content information to a certain extent while detailed textures are lost. Artflow \cite{an2021artflow} uses a flow-based network to prevent content distortion, while checkerboard artifacts exist due to the squeeze operation. CCPL \cite{wu2022ccpl} utilizes a novel transformation in replacement of AdaIN \cite{huang2017arbitrary} and achieves good stylistic results.
As shown in Figure \ref{fig:art_pic}, our results not only have great stylistic effects but also achieve the best retention of content information.\\[1ex]
\noindent{\textbf{Quantitative Evaluation}}:
Following  \cite{Hong_2021_ICCV}, we evaluate the stylized images quantitatively using SSIM and KID, where SSIM indicates the performance of content preservation, KID measures the similarity between the transferred image and the target image. As shown in Table \ref{tab:style_res}, our model achieves the best content preservation and the second-best KID score with over 5 times smaller parameters and FLOPs. \\[1ex]
\noindent{\textbf{User Study}}:
To give an additional quantitative evaluation, we conduct a user study from 119 volunteers. We randomly choose 42 content images and 26 style images from the test set to generate 1092 content-style pairs for each method. Each participant is randomly allocated 10-15 pairs and chooses the best method in \textit{Detail} (preservation of texture and semantic information) and in \textit{Overall} (overall performance, i.e., quality, stylization, fidelity) for each pair. We finally collect 1673 effective votes, and Table \ref{tab:user} shows the human performance rate, where our model outperforms previous methods by a large margin.
\begin{figure}[t]
  \centering
  \includegraphics[width=1.0\linewidth]{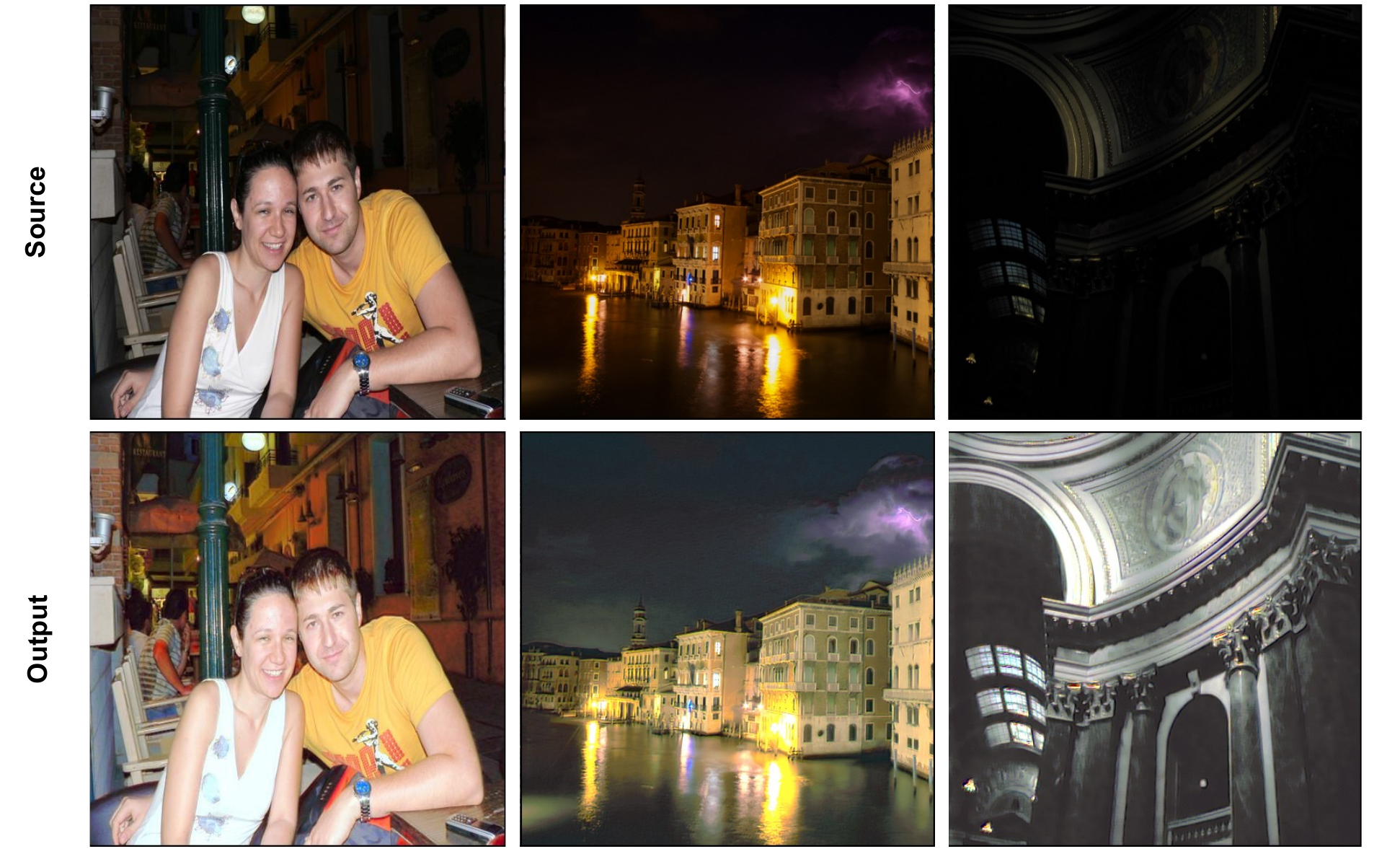}
  \caption{Visual results of low-light enhancement on the LOL dataset~\cite{wei2018deep}. Our generated images effectively retain content information and notably enhance dark areas compared to the source images.
   }
  \label{fig:enhance}
\end{figure}
\subsection{Low-level Vision [Strong-Fidelity]}
\label{sec:llv}
\begin{figure}[t]
  \centering
  \includegraphics[width=1.0\linewidth]{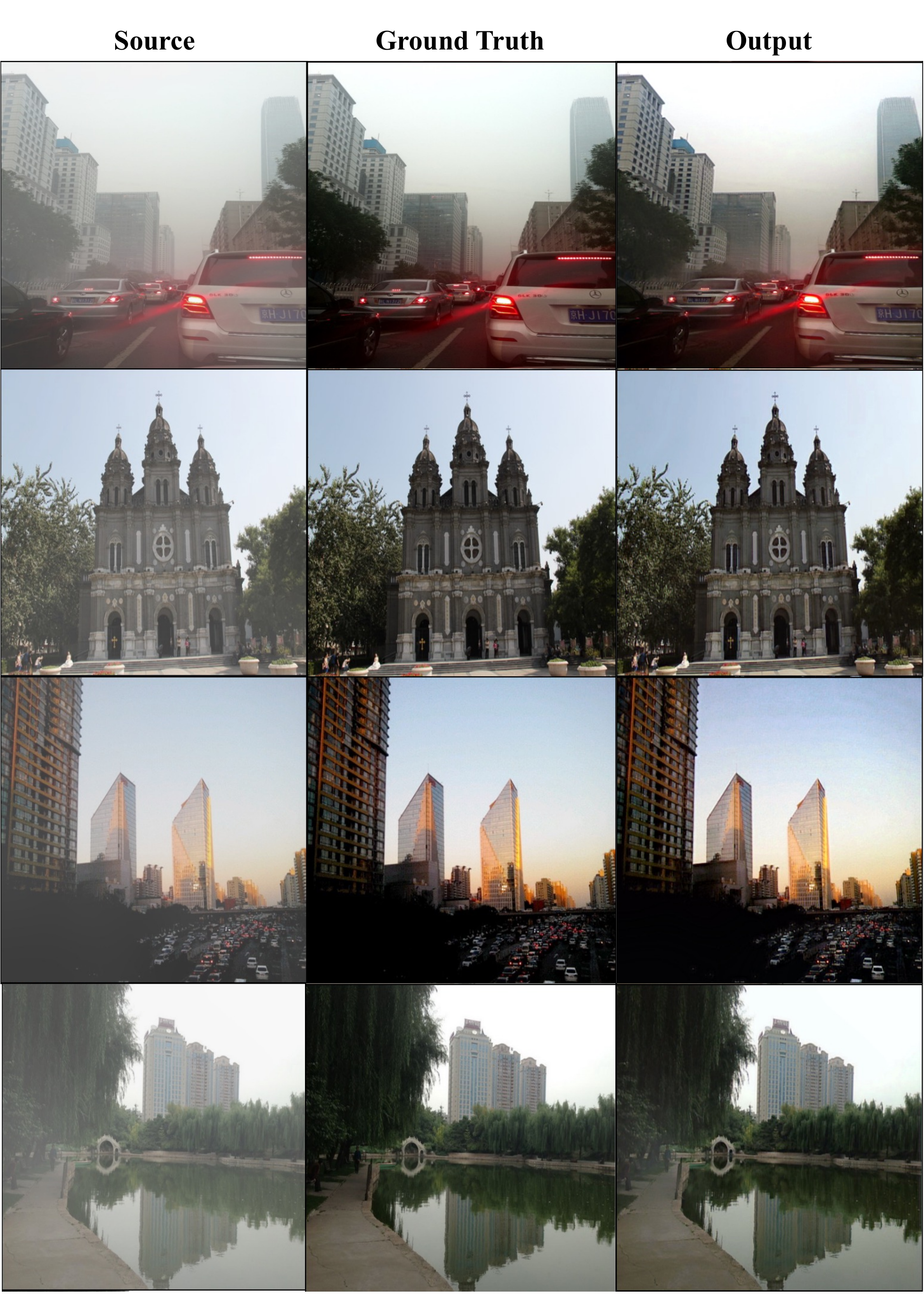}
  \caption{Visual results of Dehaze. Our method shows outstanding dehazing performance in various scenarios, with output image quality comparable to the ground-truth images.
   }
  \label{fig:dehaze_visual}
\end{figure}

We evaluated our model with two different low-level vision tasks: (1) Low-light Enhancement (2) Dehazing.\\[1.0ex]
\noindent\textbf{Dataset}: 
We conduct the low-light enhancement experiment following the official LOL dataset \cite{wei2018deep}'s train/val/test split. Dehazing experiment is trained on RESIDE-ITS \cite{li2018benchmarking} and evaluated on Synthetic Objective Testing Set (SOTS) \cite{li2018benchmarking}. By default, all images are resized to $512\times512$ for training and testing.\\[0.5ex]
\noindent\textbf{Low-light Enhancement}: 
We performed a quantitative comparison of our model and other methods on the LOL testset \cite{wei2018deep} with metric NIQE \cite{mittal2012making}. As shown in Table \ref{tab:lowlight_result}, our model achieved the best NIQE score on natural images compared to previous methods. More visual results are shown in Fig.~\ref{fig:enhance}.\\[0.5ex]
\noindent\textbf{Dehazing}: 
We compare our model with previous methods in Table \ref{tab:dehaze_result} with metrics PSNR and SSIM, our model achieved second-best PSNR and SSIM score. Fig.~\ref{fig:dehaze_visual} illustrates qualitative results of our method.
\begin{table}[!t]
\caption{Quantitative results of low-light enhancement. NIQE \cite{mittal2012making} score is used as the metric, where the smaller value indicates better perceptual performance. Our method consistently yields better results.\label{tab:lowlight_result}}
\centering
{
\begin{tabular}{c|c|c|c|c|c}
\toprule[1.2pt]
Method &  MEF$\downarrow$ & LIME$\downarrow$ & NPE$\downarrow$ & DICM$\downarrow$ & All$\downarrow$ \\
\midrule
Source Image & 4.265 & 4.438 & 4.319 & 4.255 & 4.134 \\
RetinexNet \cite{wei2018deep} & 4.149 & 4.420 & 4.485 & 4.200 & 3.920 \\
CycleGAN \cite{zhu2017unpaired} & 3.782 & 3.276 & 4.036 & 3.560 & 3.554 \\
LLNet \cite{lore2017llnet} & 4.845 & 4.940 & 4.78 & 4.809 & 4.751 \\
Jinag \textit{et al.} \cite{jiang2021enlightengan} & \textbf{3.232} & 3.719 & 4.113 &  \underline{3.570} & 3.385 \\
ArtFlow \cite{an2021artflow} & 3.621 & \underline{3.579} & \underline{3.052} & 3.578 & \underline{3.381} \\
\textbf{Ours} & \underline{3.511} & \textbf{3.418} & \textbf{3.460} & \textbf{2.916} & \textbf{3.306} \\
\bottomrule[1.2pt]
\end{tabular}}
\end{table}

\begin{table*}[!t]
\caption{ Quantitative results of
Dehaze. Our model achieves competitive results on both PSNR and SSIM metrics.\label{tab:dehaze_result}}
\centering
{
\begin{tabular}{c|c|c|c|c|c|c|c}
\toprule[1.2pt]
Method & DehazeNet \cite{cai2016dehazenet} & GMAN \cite{liu2019single} & GFN \cite{ren2018gated} & MAXIM \cite{tu2022maxim} & GCANet \cite{chen2019gated} & 
 ArtFlow \cite{an2021artflow} & \textbf{Ours} \\
\midrule
PSNR$\uparrow$ & 22.45 & 28.07 & 21.55 & \textbf{34.09} & 19.98 & 28.02 & \underline{28.25} \\
SSIM$\uparrow$ & 0.851 & 0.934 & 0.843 & \textbf{0.984} & 0.704 & 0.935 & \underline{0.945} \\
\bottomrule[1.2pt]
\end{tabular}}
\end{table*}

\subsection{Ablation Study}
\noindent \textbf{Runtime Analysis.}
On a single NVIDIA 32G V100 GPU, our method could process 86.46 samples per second in the GTA2Cityscapes task and 4.30 samples per second in the style-transfer task, which outperforms most SOTA methods in both tasks (see Table \ref{tab:runtime}). \\[0.5ex]
\noindent \textbf{Architecture Analysis.}
To demonstrate the effectiveness of our network design, we conduct an ablation study on network architecture between our HF and ArtFlow \cite{an2021artflow}, by adding the key components of ArtFlow except for the squeeze operation to our model. Table \ref{tab:ab_structure} shows the quantitative comparison of the artistic style transfer task. Compared to ArtFlow, our model achieves better performance with simpler network architecture, which shows the significance of Hierarchy Flow for I2I translation tasks.\\[1ex]
\noindent \textbf{Effect of $k$ in \textit{aligned-style loss}.}
Table \ref{tab:ab_k} shows the effect of $k$ in the \textit{aligned-style loss} for the artistic style transfer task. As designed, larger $k$ trades content preservation for stylization in translated images. The model trained with $k=0.7$ retain the richest semantic details with the best SSIM score. Aligned-style loss with $k=1.0$ is equivalent to the vanilla style loss and highly distorts semantic information and becomes over-stylized. \\[1ex]
\noindent \textbf{Effect of Model Size.}
We conduct ablation study of model size in Dehazing task. As shown in Table \ref{tab:ab_k}, base model \textit{HF} with only 0.68M parameters already achieves competitive performance. 
As the model size expands (\textit{HF+}: 0.74M; \textit{HF++}: 1.01M; \textit{$\text{HF}^\dag$}: 6.30M), performance improves consistently. \\[1ex]
\noindent \textbf{Effect of Image Resolution.}
Due to the simplicity of network design, our model is capable to support high-dimensional image training and inference. We conduct an ablation study of different input/output resolutions with model \textit{HF} and $k=0.8$ for GTA to Cityscapes tasks. As results shown in Table \ref{tab:ab_structure}, compared to the baseline in $512\times256$, mIoU obtains a performance boost by $\textbf{+2.90}$ and $\textbf{+4.28}$ with resolution of $1024\times512$ and $2048\times1024$ respectively. \\[1ex]


\begin{table}[!t]
\caption{Runtime analysis. Comparison on throughput (N samples per second) for different methods in GTA to Cityscapes (top rows) and artistic style transfer (bottom rows) respectively.\label{tab:runtime}}
\centering
\begin{tabular}{c|c|c|c|c|c}
\toprule[1.2pt]
CycleGAN & GcGAN & DRIT++ & CUT & VSAIT & \textbf{Ours} \\
\midrule
15.76 & \underline{33.15} & 21.06 & 32.39 & 32.39 & \textbf{86.46} \\
\midrule
\midrule
CCPL & ArtFlow* & 
ArtFlow\dag & AdaIN & WCT & \textbf{Ours} \\
\midrule
4.28 & 2.44 & 3.45 & \textbf{14.66} & 1.34 & \underline{4.30}\\
\bottomrule[1.2pt]
\end{tabular}
\end{table}

\begin{table}[!t]
\caption{Ablation study. 1. Ablation study of $k$ in our proposed \textit{aligned-style loss} on artistic style transfer task. Different $k$ performs trade-off between content preservation and stylization, indicated by SSIM and FID respectively. 2. Model performance vs. Model size on Dehazing task, large model boosts performance consistently.\label{tab:ab_k}}
\centering
{
\begin{tabular}{c|c|c||c|c|c|c}
\toprule[1.2pt]
$k$ &  \ FID$\downarrow$ & \ SSIM$\uparrow$ & Model & \ PSRN$\uparrow$ & \ SSIM$\uparrow$ & \ Params$\downarrow$ \\
\midrule
0.7 & 0.91 & \textbf{0.60} & HF & 26.76 & 0.931 & \textbf{0.68M}\\
0.8 & \underline{0.82} & \underline{0.56} & HF+ & 27.18 & 0.936 & \underline{0.74M} \\
0.9 & \textbf{0.50} & 0.39 & HF++ & \underline{27.45} & \underline{0.937} & 1.01M  \\
1.0 & 0.95 & 0.17 & ${\text{HF}}^\dag$ & \textbf{28.25} & \textbf{0.945} & 6.30M  \\
\bottomrule[1.2pt]
\end{tabular}}
\end{table}

\begin{table}[!t]
\caption{Ablation study of network architecture between HF and ArtFlow for artistic style transfer task. Overall, Hierarchy Flow achieves the best results. \label{tab:ab_structure}}
\centering
{
\begin{tabular}{c|c|c|c}
\toprule[1.2pt]
Architecture &  FID$\downarrow$ & SSIM$\uparrow$ & KID($\times 10^3$)$\downarrow$ \\
\midrule
\textbf{Hierachy Flow} & \textbf{0.61} & \textbf{0.55} & \textbf{25.0/5.1} \\
+ Actnorm &  0.88 & \underline{0.29} & 27.3/4.9 \\
+ 1x1 Conv &  0.97 & 0.24 & \underline{25.3/4.2} \\
ArtFlow & \underline{0.85} & 0.21 & 27.2/4.5\\
\bottomrule[1.2pt]
\end{tabular}}
\end{table}

\section{Conclusion}
\noindent In this paper, we categorize image-to-image translation problems into three levels: strong-, normal-, and weak-fidelity translation. We proposed a novel invertible network Hierarchy Flow, with a dedicated aligned-style loss for high-fidelity image-to-image translation. Qualitative and quantitative results show that our model obtains better content preservation during translation, and achieves the best performance in high-fidelity translation tasks.

\noindent \textbf{Future work.} Although our model outperforms previous methods  in strongly and normally constrained tasks, we failed to achieve admirable results in all weakly constrained translation tasks. Future work includes extending this model to full spectrum of image-to-image translation tasks.

\noindent \textbf{Broader Impact.}
Our proposed generative model could eliminate the gap between simulation and reality, which can be widely used in self-driving and medical areas. 
The use of image synthesis would not lead to privacy issues but might create fake news, thus more regulations are needed to restrict the usage of synthesized data.

\begin{table}[!t]
\caption{GTA to Cityscapes with different image resolutions. Due to the simple design of network, HF supports HD image training and testing, which yields better performance.\label{tab:ab_resulotion}}
\centering
\begin{tabular}{c|c|c|c}
\toprule[1.2pt]
resolution & \ pxAcc$\uparrow$ & \ clsAcc$\uparrow$ & \ mIoU$\uparrow$ \\
\midrule
\textbf{\scriptsize{512x256}} & 81.04 & 41.92 & 31.52 \\
\textbf{\scriptsize{1024x512}} & \underline{84.95} & \underline{45.25} & \underline{34.42} \\
\textbf{\scriptsize{2048x1024}} & \textbf{87.21} & \textbf{46.74} & \textbf{35.80} \\
\bottomrule[1.2pt]
\end{tabular}
\end{table}

{\small
\bibliographystyle{IEEEtran}
\bibliography{egbib}
}




\ifCLASSOPTIONcaptionsoff
  \newpage
\fi

%



\vfill
\newpage
\begin{IEEEbiography}
[{\includegraphics[width=1in,height=1.25in,clip,keepaspectratio]{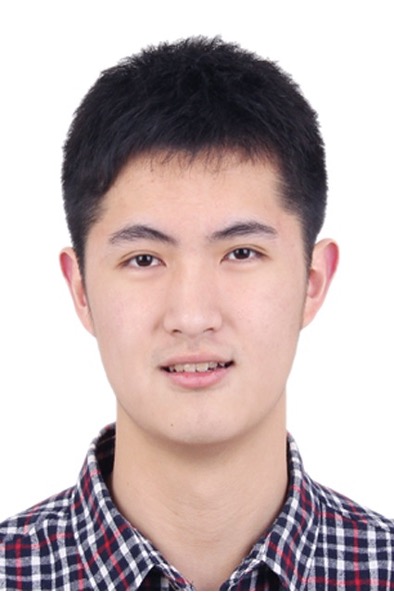}}]{Weichen Fan} received his B.S. degree in electronic science and engineering (ESE) from the University of Electronic Science and Technology of China (UESTC) and his master's degree from the Department of Electronic and Computer Engineering (ECE), National University of Singapore (NUS). His is currently a researcher at SenseTime. His research interests include low-level vision, transfer learning, and multi-modal learning. 
\end{IEEEbiography}
\begin{IEEEbiography}
[{\includegraphics[width=1in,height=1.25in,clip,keepaspectratio]{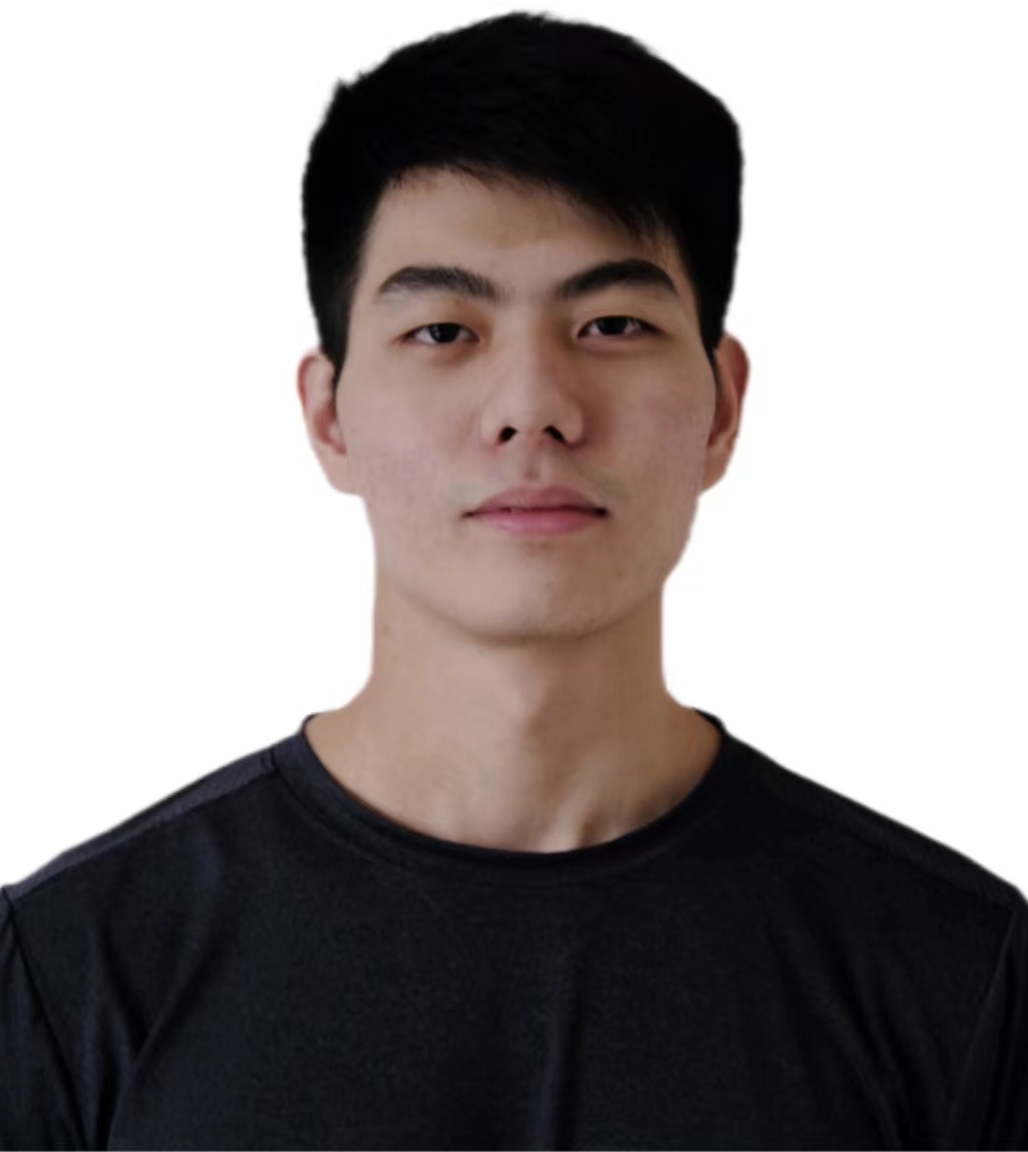}}]{Jinghuan Chen} received his B.Eng. degree from Nanyang Technological University, Singapore in 2020. He is currently an Algorithm Engineer at ByteDance Inc. His research interests include computer vision, generative models and multi-modal learning.
\end{IEEEbiography}
\begin{IEEEbiography}
[{\includegraphics[width=1in,height=1.25in,clip,keepaspectratio]{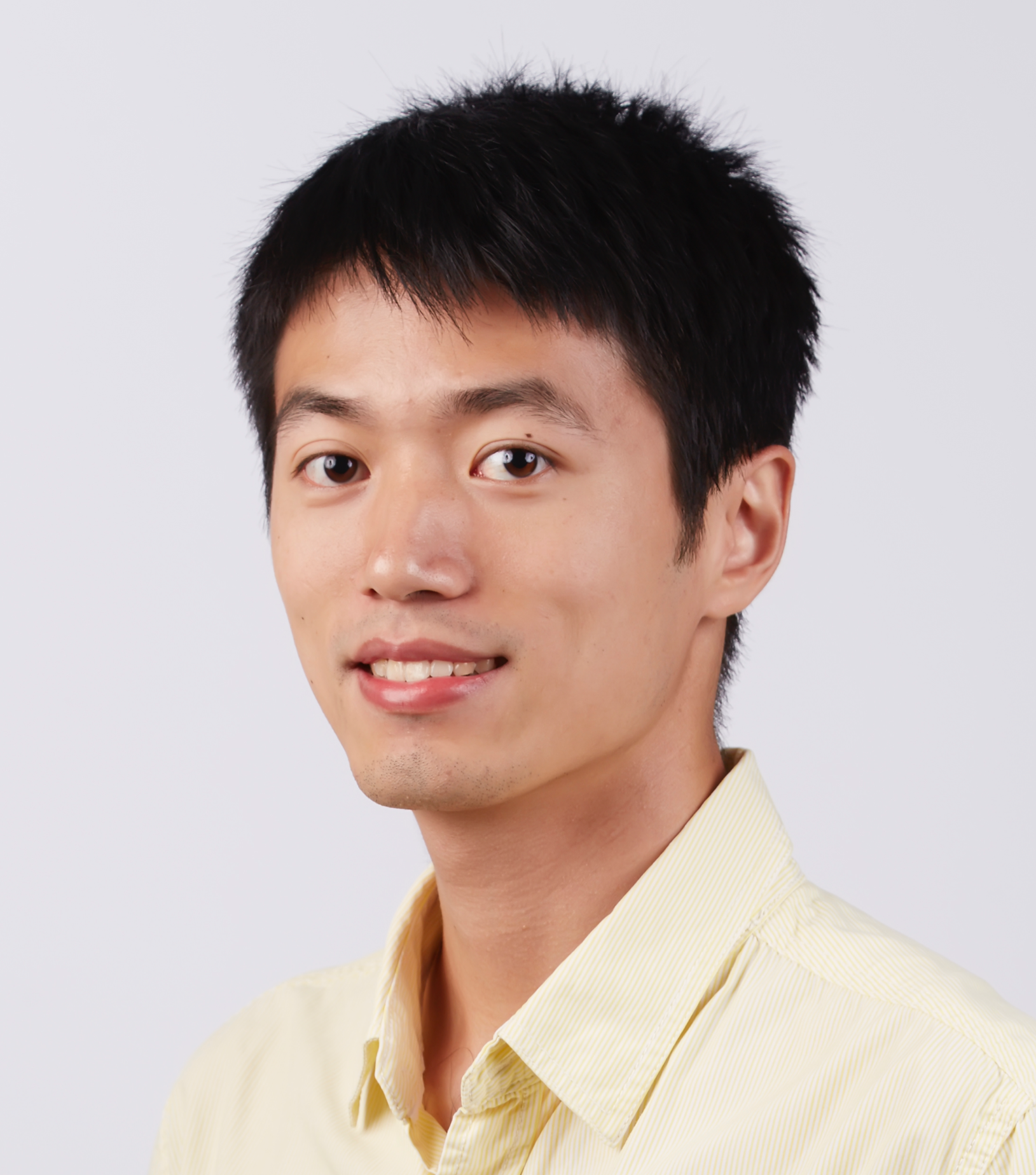}}]{Ziwei Liu} is currently a Nanyang Assistant Professor at Nanyang Technological University, Singapore. His research revolves around computer vision, machine learning and computer graphics. He has published extensively on top-tier conferences and journals in relevant fields, including CVPR, ICCV, ECCV, NeurIPS, ICLR, ICML, TPAMI, TOG and Nature - Machine Intelligence. He is the recipient of Microsoft Young Fellowship, Hong Kong PhD Fellowship, ICCV Young Researcher Award, HKSTP Best Paper Award and WAIC Yunfan Award. He serves as an Area Chair of CVPR, ICCV, NeurIPS and ICLR, as well as an Associate Editor of IJCV.
\end{IEEEbiography}


\vfill


\end{document}